\documentclass[conference]{IEEEtran}
\IEEEoverridecommandlockouts
\usepackage{cite}
\usepackage{booktabs}
\usepackage{amsmath,amssymb,amsfonts}
\usepackage{algorithmic}
\usepackage{graphicx}
\usepackage{textcomp}
\usepackage{xcolor}
\usepackage{subfig}
\def\BibTeX{{\rm B\kern-.05em{\sc i\kern-.025em b}\kern-.08em
    T\kern-.1667em\lower.7ex\hbox{E}\kern-.125emX}}
\begin{document}

\title{GGNNs : Generalizing GNNs using Residual Connections and Weighted Message Passing\\
}

\author{\IEEEauthorblockN{Abhinav Raghuvanshi}
\IEEEauthorblockA{\textit{Department of Aerospace Engineering} \\
\textit{Indian Institute Of Technology, Bombay}\\
200040008@iitb.ac.in}
\and
\IEEEauthorblockN{Kushal S M}
\IEEEauthorblockA{\textit{Department of Aerospace Engineering} \\
\textit{Indian Institute Of Technology, Bombay}\\
200010078@iitb.ac.in}
}

\maketitle

\begin{abstract}
Many real-world phenomena can be modeled as a
graph, making them extremely valuable due to their ubiquitous
presence. GNNs excel at capturing those relationships and pat-
terns within these graphs, enabling effective learning and prediction tasks. GNNs are constructed using Multi-Layer Perceptrons
(MLPs) and incorporate additional layers for message passing
to facilitate the flow of features among nodes. It is commonly
believed that the generalizing power of GNNs is attributed to
the message-passing mechanism between layers, where nodes
exchange information with their neighbors, enabling them to
effectively capture and propagate information across the nodes of
a graph. Our technique builds on these results, modifying
the message-passing mechanism further: one by
weighing the messages before accumulating at each node and
another by adding Residual connections. These two mechanisms
show significant improvements in learning and faster
convergence.

\end{abstract}


\section{Introduction}
Graph Neural Networks (GNNs) have emerged as a robust
methodology for modeling and analyzing various real-world
phenomena.
Their ability to capture intricate relationships and patterns
within graph structures has rendered them invaluable in numerous learning and prediction tasks. Typically constructed using multi-layer Perceptrons (MLPs), GNNs employ layered
architectures that facilitate message passing, enabling effective
information propagation across nodes in a graph.
The prevailing belief attributes the generalization prowess of
GNNs primarily to their message-passing mechanism, wherein
nodes exchange information with their neighbors, thereby
enabling comprehensive information flow across the graph.
Recent research delves deeper into enhancing learning by incorporating message passing between Multi-Layer Perceptron (MLP) layers.
In this pursuit, we introduce innovative modifications aimed
at refining the message-passing mechanism further.
Our proposed methodology extends previous findings presented by Yang et al.\cite{yang2020graph}, wherein they postulate that the success of GNNs in node-level prediction tasks is not solely a consequence of their advanced expressivity but is deeply rooted in their intrinsic generalization capability.
Expanding upon these insights, our approach builds upon
existing methodologies by introducing two key enhancements:
first, by integrating weighted message passing, allowing for
adaptive message weighting based on the relevance of in-
formation; and second, by introducing residual connections,
preserving and incorporating information from prior layers to
expedite convergence and enhance learning.
Residual connections, initially popularized in convolutional
neural networks (CNNs) provide a shortcut path for information to flow through the network by bypassing certain layers. 
This addition has shown promising results in enabling
smoother gradient flow and facilitating the learning process,
especially in deep architectures\cite{residual}.

Through empirical evaluations across various datasets and
comparative analyses with existing models such as GNNs and
PMLPs\cite{yang2020graph}, we showcase the efficacy of our proposed en-
enhancements in terms of improved learning, faster convergence,
and superior performance in node-level prediction tasks.
In this paper, we delineate the conceptual framework for
emulation, and experimental results that underscore the
effectiveness of our proposed modifications in enhancing the
capabilities of GNNs. Our findings contribute to advancing
the understanding and applicability of graph neural networks
in diverse real-world scenarios.

\section{Related Work}
Our research extends the findings presented in the paper \emph{Graph Neural Networks are Inherently Good Generalizers: Insights by Bridging GNNs and MLPs" by Chenxiao Yang, Qitian Wu, Jiahua Wang, and Junchi Yan} \cite{yang2020graph}. The authors contend that the effectiveness of Graph Neural Networks (GNNs) in node-level prediction tasks do not solely arise from their advanced expressivity, but rather from their inherent generalization capability.

The authors introduce an intermediate model class called P (Propagational)MLP, which is identical to standard MLP in training, but then adopting GNN's architecture in testing. They observe that PMLPs consistently perform on par with (or even exceed) their GNN counterparts while being much more efficient in training. 

This paper argues that the success of GNNs in node-level prediction tasks is not solely due to their advanced expressivity, but rather their intrinsic generalization capability. 

\begin{equation}
h_u = \sigma \left( \sum_{v \in \mathcal{N}(u)} W h_v + b \right)
\end{equation}

This equation represents the message-passing operation for a node $u$ in a graph. The variable $h_u$ is the feature vector of node $u$, $\sigma$ is an activation function, $\mathcal{N}(u)$ is the set of neighbors of node $u$, $W$ is a weight matrix, and $b$ is a bias vector.

\begin{equation}
z_u = \phi \left( h_u \odot g_u \right)
\end{equation}

The variable $z_u$ is the fused feature vector of node $u$, $\phi$ is an activation function, $h_u$ is the feature vector of node $u$ from the previous equation, $\odot$ is the element-wise product, and $g_u$ is a gating vector of node $u$.

\begin{equation}
y_u = \psi \left( z_u \odot f_u \right)
\end{equation}

The node classification operation for a node $u$ in a graph. The variable $y_u$ is the output vector of node $u$, $\psi$ is an activation function, $z_u$ is the fused feature vector of node $u$ from the previous equation, $\odot$ is the element-wise product, and $f_u$ is a feature vector of node $u$ from the input graph.

In summary, the authors of this paper propose a new model class, PMLP, that combines the advantages of GNNs and MLPs for graph node-level prediction tasks. They show that PMLP can achieve similar or better performance than GNNs at training and testing time, and explain the learning behavior and extrapolation behavior of GNNs by the NTK feature map. This paper provides new insights into learning behavior of GNNs and can be used as an analytic tool for dissecting various GNN-related research problems.

\section{Background and Setup}
Consider a graph dataset \( G = (V, E) \) where the node set \( V \) contains \( n \) node instances \( \{(x_u, y_u)\}_{u\in V} \), where \( x_u \in \mathbb{R}^d \) denotes node features, and \( y_u \) is the label. Without loss of generality, \( y_u \) can be a categorical variable or a continuous one, depending on specific prediction tasks (classification or regression). Instance relations are described by the edge set \( E \) and an associated adjacency matrix \( A \in \{0, 1\}^{n\times n} \). In general, the problem is to learn a predictor model with \( \hat{y} = f(x; \theta, G_k^x) \) for node-level prediction, where \( G_k^x \) denotes the \( k \)-hop ego-graph around \( x \) over \( G \).

In the upcoming subsections, we introduce two innovative approaches that involve adjustments to the message-passing and feed-forward architectures. These modifications aim to enhance the generalizability of GNNs. Our findings demonstrate that these methods outperform standard GNNs and surpass the performance of PMLP \cite{yang2020graph}. Additional results from our experiments are detailed in the \ref{sec:Results} Section.
\begin{figure}[htbp]
    \centering
    \includegraphics[width=0.3\textwidth]{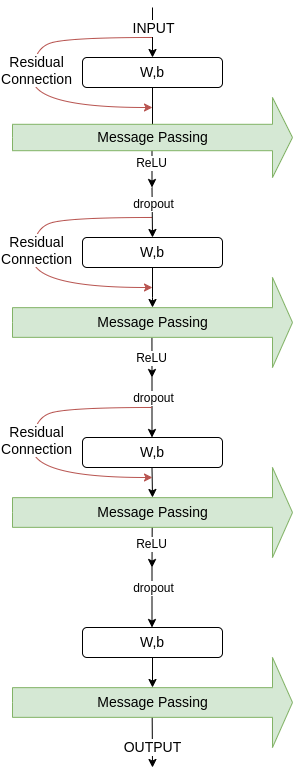} 
    \caption{Figure shows GNN architecture with Residual Connections}
    \label{fig:1}%
\end{figure}

\subsection{Residual Connections}
The essence of the residual connection mechanism lies in preserving information from the preceding layer while incorporating the current layer's transformation, which essentially represents the affine output of the prior layer. This integration involves a residual connection $\mathbf{h}_u^{(l-1)}$ alongside the transformed output $\hat{\mathbf{h}}_u^{(l)}$ at node $u$, formulated as follows:

\begin{equation}
\mathbf{h}_u^{(l-1)} = \text{MaxPool}(\mathbf{h}_u^{(l-1)})
\end{equation}
\begin{equation}
\mathbf{h}_u^{(l)} = \hat{\mathbf{h}}_u^{(l)} + \alpha(\mathbf{h}_u^{(l-1)})
\end{equation}

Here, $\alpha$ symbolizes the scaling factor, and $\hat{\mathbf{h}}_u^{(l)}$ denotes the transformed output from the primary pathway.

In various affine connections, alterations in feature lengths often occur. To seamlessly incorporate scaled features from the previous layer into the current layer, it's imperative to align their feature lengths. In our context, we employ MaxPool to compress the previous layer's features to match the current layer's feature length.

The $\hat{\mathbf{h}}_u^{(l)}$ undergoes Message Passing before the activation function (ReLU), as depicted in Fig~\ref{fig:1}.

The integration of shortcut connections, illustrated in Eqn. (4), does not impose additional parameters or computational complexity. This characteristic significantly contributes to practicality, especially in our comparative analysis between plain GNNs and residual connected GNNs. It ensures that both architectures maintain identical parameters, depth, and width.

\subsection{Learning to aggregate message}
Consider a general GNN layer with a message aggregation function that takes into account the messages from neighboring nodes. The aggregation at node \( u \) is given by:

\[
\hat{h}_u^{(l-1)} = \sum_{v \in N_u \cup\{{u}\}} w_{uv} \cdot h_v^{(l-1)}
\]

Here, \( N_u \) is the set of neighboring nodes centered at \( u \), \( w_{uv} \) represents the learnable weight parameter associated with the message from node \( v \) to node \( u \) at layer \( l \) and \( h_v^{(l-1)} \) is the representation of node \( v \) from the previous layer.

Mathematically, this can be expressed as:

\[
\hat{h}_u^{(l)} =  \sigma(\psi^{(l)}(h_u^{(l-1)}))
\]

where \( \sigma \) is an activation function (ReLU) and \( \psi^{(l)} \) denotes the feed-forward process (linear feature transformation followed by a non-linear activation) at the \( l \)-th layer.

The learnable weight $w_{uv}$ is taken to be 
\[
w_{uv} = \frac{1}{\sqrt{d_u \cdot d_v}} \cdot \frac{1}{1 + e^{-\theta_{uv}}}
\]
where we associate $\theta_{uv}$ with each edge in the graph to be the learnable parameter, $d_u$ \& $d_v$ are the degrees of nodes $u$ \& $v$, respectively. We optimize it using gradient descent on the same loss as every other parameter in the network. The method is shown in Fig 2.

The motivation for introducing learnable weights is to allow the model to assign different importance to messages from different neighboring nodes during the aggregation process. Not all nodes contribute equally to the information exchange, and by learning these weights, the model gains flexibility in adjusting the significance of each message based on the task or data characteristics.

\begin{figure}[htbp]
    \centering
    \includegraphics[width=0.3\textwidth]{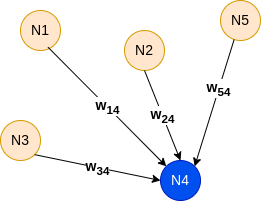} 
    \caption{Figure shows weighted Message Passing}
    \label{fig:2}%
\end{figure}

Introducing \(\theta_{uv}\) for each edge leads to an increase in the number of parameters proportional to the number of edges in the graph, allowing the model to capture edge-specific patterns and relationships.

As an example, a representative matrix for \(\theta_{uv}\) could be visualized as follows:

\[ \Theta = \begin{bmatrix} \theta_{1,1} & \theta_{1,2} & \ldots & \theta_{1,|E|} \\ \theta_{2,1} & \theta_{2,2} & \ldots & \theta_{2,|E|} \\ \vdots & \vdots & \ddots & \vdots \\ \theta_{|E|,1} & \theta_{|E|,2} & \ldots & \theta_{|E|,|E|} \end{bmatrix} \]

Here, \(|E|\) is the number of edges in the graph, and each \(\theta_{uv}\) is a scalar parameter associated with the edge between nodes \(u\) and \(v\).

This approach enables the GNN to adaptively assign relevance to the incoming messages, giving it the capacity to capture intricate relationships within the graph structure. It adds a layer of expressiveness to the model, allowing it to focus on more influential nodes for the specific learning task.

We operate a regular training routine to train the network, apart from the internal changes that we make in message passing and feed forwarding etc- 
\begin{itemize}
  \item \textbf{Dataset Loading and Preprocessing:}
  We load and preprocesses a graph dataset (taken in as parameter), including features, labels, and adjacency matrix. Additional preprocessing steps, such as symmetrizing the adjacency matrix, are performed for certain datasets.

  \item \textbf{Model Initialization:}
  The GGNN model is initialized based on the specified method and dataset parameters.
  Hyper Parameters used are 
\begin{center}
\begin{tabular}{|c|c|c|}
\hline
Parameter & Residual GGNN & Learnable GGNN \\
\hline
Learning Rate & 0.1 & 0.1\\
Weight Decay & 0.01 & 0.1\\
Dropout & 0.5 & 0.1\\
Hidden Channels & 64 & 64\\
\hline
\end{tabular}
\end{center}

  \item \textbf{Loss Function and Optimizer:}
  The code sets up a loss function based on the dataset characteristics, such as binary or multiclass classification. The optimizer (Adam) is configured with a specified learning rate and weight decay.

  \item \textbf{Evaluation}
  Evaluation metrics, such as accuracy, ROC AUC, or F1 score, are calculated on validation and test sets. Results and statistics are logged for each run, including performance metrics and training time.

\end{itemize}
\begin{table*}[]
\centering

\begin{tabular}{@{}ccccccc@{}}
\toprule
\multicolumn{1}{|l|}{}                    & \multicolumn{2}{c|}{\textbf{GNN}}                                               & \multicolumn{2}{c|}{\textbf{PMLP }}                            & \multicolumn{2}{c|}{\textbf{Learnable GGNN }}                                       \\ \midrule
\multicolumn{1}{|l|}{}                  & \multicolumn{2}{c|}{NUM\_EPOCHS = 100}                                         & \multicolumn{2}{c|}{NUM\_EPOCHS = 100}                                & \multicolumn{2}{c|}{NUM\_EPOCHS = 50}                                          \\ \midrule
\multicolumn{1}{|c|}{\textbf{Dataset}}  & \multicolumn{1}{c|}{Train}                 & \multicolumn{1}{c|}{Test}         & \multicolumn{1}{c|}{Train}        & \multicolumn{1}{c|}{Test}         & \multicolumn{1}{c|}{Train}        & \multicolumn{1}{c|}{Test}                  \\ \midrule
\multicolumn{1}{|c|}{\textbf{Cora}}     & \multicolumn{1}{c|}{\textbf{99.57 ± 0.39}} & \multicolumn{1}{c|}{50.36 ± 2.75} & \multicolumn{1}{c|}{97.71 ± 0.32} & \multicolumn{1}{c|}{73.84 ± 0.86} & \multicolumn{1}{c|}{98.43 ± 0.93} & \multicolumn{1}{c|}{\textbf{74.74 ± 0.50}} \\ \midrule
\multicolumn{1}{|c|}{\textbf{Citeseer}} & \multicolumn{1}{c|}{\textbf{99.17 ± 1.18}} & \multicolumn{1}{c|}{52.24 ± 1.50} & \multicolumn{1}{c|}{94.83 ± 0.91} & \multicolumn{1}{c|}{66.74 ± 1.23} & \multicolumn{1}{c|}{95.83 ± 0.59} & \multicolumn{1}{c|}{\textbf{67.52 ± 0.87}} \\
 \bottomrule 
\end{tabular}
\caption{Mean and STD comparison between GNN, PMLP and Learnable GGNN}
\end{table*}

\begin{table*}[]
\centering
\begin{tabular}{@{}|c|cc|cc|ccc|@{}}
\toprule
\multicolumn{1}{|l|}{} & \multicolumn{2}{c|}{\textbf{GNN}}                         & \multicolumn{2}{c|}{\textbf{PMLP}}               & \multicolumn{3}{c|}{\textbf{Residual GGNN}}                                                                      \\ \midrule
\multicolumn{1}{|l|}{} & \multicolumn{2}{c|}{NUM\_EPOCHS = 100}                    & \multicolumn{2}{c|}{NUM\_EPOCHS = 100}           & \multicolumn{3}{c|}{NUM\_EPOCHS = 50}                                                           \\ \midrule
\textbf{Dataset}       & \multicolumn{1}{c|}{Train}                 & Test         & \multicolumn{1}{c|}{Train}        & Test         & \multicolumn{1}{c|}{Scaling Factor} & \multicolumn{1}{c|}{Train}        & Test                  \\ \midrule
\textbf{Cora}          & \multicolumn{1}{c|}{\textbf{99.57 ± 0.39}} & 50.36 ± 2.75 & \multicolumn{1}{c|}{97.71 ± 0.32} & 73.84 ± 0.86 & \multicolumn{1}{c|}{0.1}           & \multicolumn{1}{c|}{98.14 ± 0.93} & \textbf{75.10 ± 1.17} \\ \midrule
\textbf{Citeseer}      & \multicolumn{1}{c|}{\textbf{99.17 ± 1.18}} & 52.24 ± 1.50 & \multicolumn{1}{c|}{94.83 ± 0.91} & 66.74 ± 1.23 & \multicolumn{1}{c|}{0.13}           & \multicolumn{1}{c|}{95.50 ± 0.95} & \textbf{67.26 ± 1.11} \\                     \bottomrule
\end{tabular}
\caption{Mean and STD comparison between GNN, PMLP and Residual GGNN}
\end{table*}

\begin{figure}[htbp]
    \centering
        \includegraphics[width=0.48\textwidth]{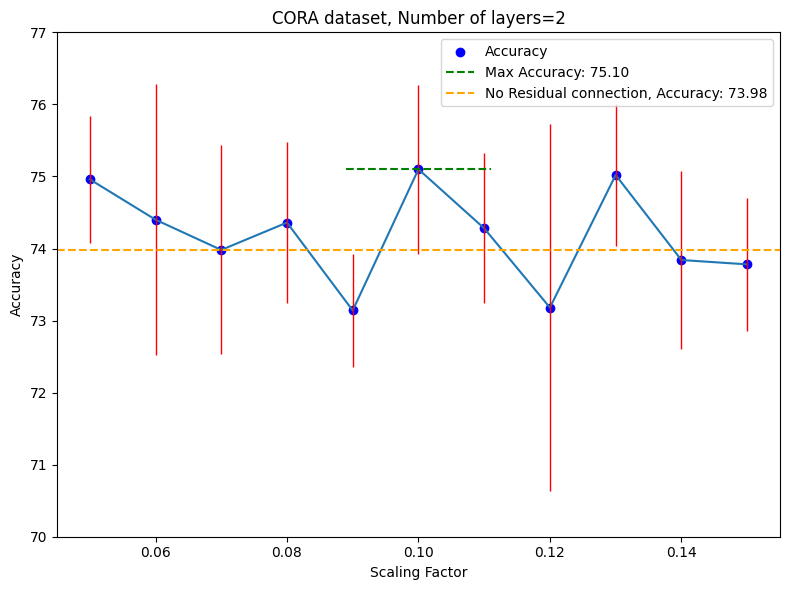} 
    \caption{Performance variation with Scaling Factor for 2-layer Residual GGNN on the Cora Dataset}
    \label{fig:example}%
\end{figure}
\begin{figure}[htbp]
    \centering
        \includegraphics[width=0.48\textwidth]{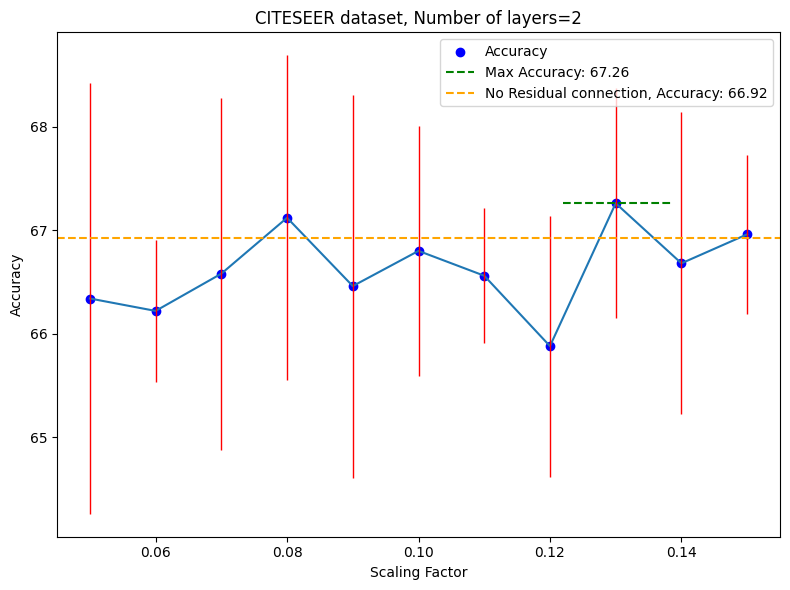} 
    \caption{Performance variation with Scaling Factor for 2-layer Residual GGNN on the Citeseer Dataset}
    \label{fig:example}%
\end{figure}
\begin{figure}[htbp]
    \centering
        \includegraphics[width=0.48\textwidth]{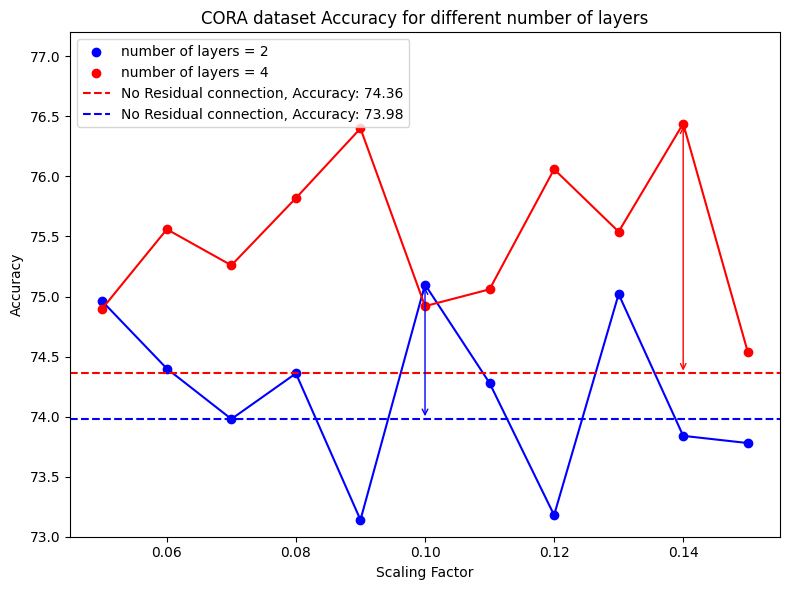} 
    \caption{Performance comparison varying with Scaling Factor and number of layers of Residual GGNN on the Cora Dataset}
    \label{fig:example}%
\end{figure}

\begin{figure}[htbp]
    \centering
        \includegraphics[width=0.48\textwidth]{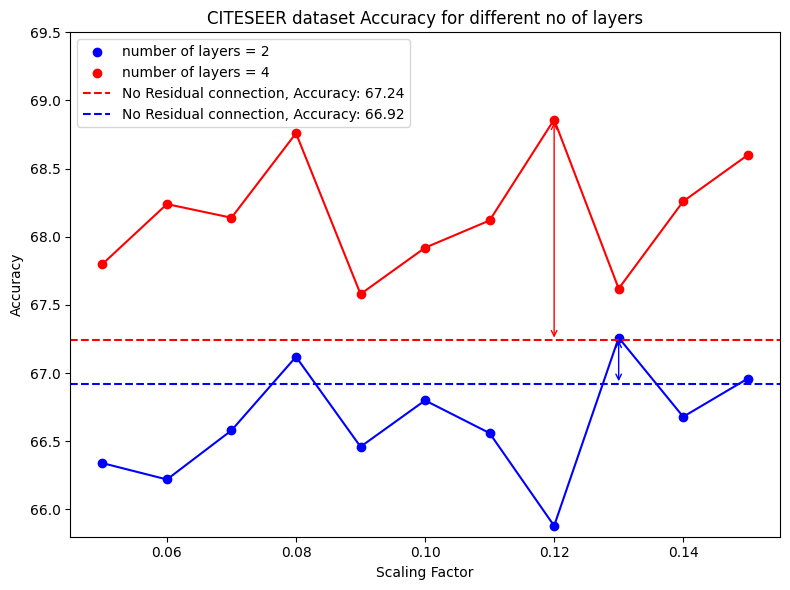} 
    \caption{Performance comparison varying with Scaling Factor and number of layers of Residual GGNN on the Citeseer Dataset}
    \label{fig:example}%
\end{figure}

\section{Results} \label{sec:Results}
We report better test accuracies for both our techniques. Tables 1 \& 2 show and compare our methods with preexisting ones.
\subsection{Residual GGNN}
We compare the performance of a residual GNN with a non-residual GNN on the CORA and Citeseer datasets with a scaling factor of 0.1 for 2-layer networks. The results are shown in Figure 3 \& 4. We observe that the residual GNN outperforms the non-residual GNN on both datasets, achieving higher accuracy scores. The residual GNN achieves a maximum accuracy of 75.10\% on the CORA dataset and 67.26\% on the Citeseer dataset, while the non-residual GNN achieves 73.98\% and 66.92\%, respectively. This shows that the residual GNN can effectively learn the graph structure and node features, and benefit from the residual connections.

We analyze the performance of Residual GGNN on the CORA dataset with different number of layers and scaling factors. The results are shown in Figure 5 and Figure 6. We observe that the Residual GGNN achieves the highest accuracy of 75.10\% with two layers and a scaling factor of 0.1. The accuracy decreases slightly as the scaling factor increases, but it remains higher than the non-residual GNN. The non-residual GNN achieves a maximum accuracy of 73.84\% with two layers. The accuracy drops significantly as the number of layers increases, indicating that the non-residual GNN suffers from the vanishing gradient problem. The residual GGNN, on the other hand, maintains a stable accuracy with more layers, showing the benefit of the residual connections.

\subsection{Learnable GGNN}
We compare the performance of three different machine learning models on the Cora and Citeseer datasets: GNN, PMLP, and Learnable GGNN. The results are shown in Table 1. We observe that the Learnable GNN, which learns the weights of each message before message aggregation, achieves the highest test accuracy on both datasets. The Learnable GGNN achieves 74.74\% ± 0.50 on the Cora dataset and 67.52\% ± 0.87 on the Citeseer dataset, while the GNN achieves 50.36\% and 52.24\%, and the PMLP achieves 73.84\% and 66.74\%, respectively. This shows that the Learnable GNN can effectively capture the importance of each message and improve the graph representation.

\section{Future Work}
In future we plan to incorporate Positonal Embeddings, the "Attention is All You Need" paper \cite{attention} introduces a technique that enhances the weight matrix's understanding of the layer to which the messages belong. This is achieved through the incorporation of positional embeddings, a concept that enriches the model's ability to capture sequential or positional information within a sequence. In the context of graph neural networks (GNNs), this method contributes to learning a more nuanced and context-aware representation of the graph structure.

Positional embeddings are computed using sine and cosine functions and are added to the node features. The formulation for positional embeddings is given by:

\[
\text{PE}(i, 2j) = \sin \left( \frac{i}{10000^{2j/d}} \right)
\]

\[
\text{PE}(i, 2j+1) = \cos \left( \frac{i}{10000^{2j/d}} \right)
\]

Here, \(i\) represents the position, \(j\) is the dimension, and \(d\) is the embedding size. The use of both sine and cosine functions ensures that the positional embeddings capture different aspects of the positional information. These embeddings are then combined with the original node features, enriching the representation with positional context.

The augmented node features, now enriched with positional embeddings, undergo a self-attention mechanism. This mechanism enables the model to weigh the importance of different
nodes based on their inherent features and positions within the graph structure. In simpler terms, it allows the model to consider the sequential order or position of nodes when learning the overall graph representation.

By incorporating positional embeddings and leveraging self-attention, the model could gain the ability to discern the relative importance and relationships between nodes in the graph based on their positions. This approach contributes to the model's capacity to capture nuanced patterns and dependencies within the graph, enhancing its performance on tasks that involve sequential or positional information.

\section{Conclusion}
In this study, we introduce a novel graph neural network architecture named GGNNs (Generalised Graph Neural Networks) that enhances the message-passing mechanism by introducing learnable weight parameters for each incoming message from neighbouring nodes. This approach acknowledges that not every message holds equal importance and by incorporating these learnable weights, GGNNs gain the capability to adaptively assign relevance to incoming messages during aggregation. Experimental results across diverse datasets and experimental settings consistently showcase the superior performance of GGNNs, surpassing both traditional MLPs and GNNs.

Building upon the success of GGNNs, we further explore another innovative technique involving incorporating residual connections. Remarkably, this technique enhances the accuracy of GGNNs, PMLP (Perceptual Message Passing MLP), and traditional GNNs without the introduction of additional parameters. This improvement demonstrates the efficacy of residual connections in capturing complex relationships within graph-structured data, showcasing their ability to enhance accuracy without compromising model simplicity. We achieve higher accuracies on different datasets, compared with the same set of hyperparameters for GNNs and PMLP.

\vspace{12pt}

\end{document}